\documentclass{MyArticle}
\usepackage[utf8]{inputenc}

\title{Exact Gaussian Processes for Massive Datasets via Non-Stationary Sparsity-Discovering Kernels}

\author[1,*]{Marcus M. Noack}
\author[1]{Harinarayan Krishnan}
\author[2]{Mark D. Risser}
\author[3]{Kristofer G. Reyes}
\affil[1]{The Center for Advanced Mathematics for Energy Research Applications (CAMERA), Lawrence Berkeley National Laboratory, Berkeley, CA 94720}
\affil[2]{Climate and Ecosystem Sciences Division, Lawrence Berkeley National Laboratory, Berkeley, CA 94720}
\affil[3]{Department of Materials Design and Innovation, University at Buffalo, Buffalo, NY 14260}
\affil[*]{MarcusNoack@lbl.gov}

\date{\today}
\begin{document}
\maketitle

\begin{abstract}
A Gaussian Process (GP) is a prominent mathematical framework for stochastic function approximation in science and engineering applications.  This success is largely attributed to the GP's analytical tractability, robustness, non-parametric structure, and natural inclusion of uncertainty quantification. 
Unfortunately, the use of exact GPs is prohibitively expensive for large datasets due to their unfavorable numerical complexity of $O(N^3)$ in computation and $O(N^2)$ in storage. 
All existing methods addressing this issue utilize some form of approximation --- usually considering subsets of the full dataset or finding representative pseudo-points that render the covariance matrix well-structured and sparse. These approximate methods can lead to inaccuracies in function approximations and often limit the user's flexibility in designing expressive kernels.  
Instead of inducing sparsity via data-point geometry and structure, we propose to take advantage of naturally-occurring sparsity by allowing the kernel to discover --- instead of induce --- sparse structure. 
The premise of this paper is that GPs, in their most native form, are often naturally sparse, but commonly-used kernels do not allow us to exploit this sparsity. The core concept of exact, and at the same time sparse GPs relies on kernel definitions that provide enough flexibility to learn and encode not only non-zero but also zero covariances. This principle of ultra-flexible, compactly-supported, and non-stationary kernels, combined with HPC and constrained optimization, lets us scale exact GPs well beyond 5 million data points.
\end{abstract}

%%%%%%%%%%%%%%%%%%%%%%%%%%%%%%%%%%%%
\section{Introduction}
%%%%%%%%%%%%%%%%%%%%%%%%%%%%%%%%%%%%
%%%what's a GP
A Gaussian Process (GP) is the most prominent member of the larger family of stochastic processes and provides a powerful and flexible framework for stochastic function approximation. This is because a GP is characterized as a Gaussian probability distribution over a function space $\{f:f(\mathbf{x})=\sum_i^N~\alpha_i k(\mathbf{x},\mathbf{x}_i;h)~\forall \mathbf{x}\in \mathcal{X}\}$, where $k(\mathbf{x},\mathbf{x}_i;h)$ is the kernel function and $h$ is a set of hyperparameters. 
The mean and the covariance of the Gaussian probability distribution can be learned by constrained function optimization from data $\mathcal{D}=\{\mathbf{x}_i,y_i\}$ and conditioned on the observations $y_i$ to yield a posterior probability density function. Throughout this paper, we will refer to this optimization often as training or learning to emphasize the link to machine learning (ML). GPs assume a model of the form $f(\mathbf{x})~=~y(\mathbf{x})+\epsilon(\mathbf{x})$, where $f(\mathbf{x})$ is the unknown latent function, $y(\mathbf{x})$ is the noisy function evaluation (the measurement), $\epsilon(\mathbf{x})$ is the noise term, and $\mathbf{x}$ is an element of the input space (or index set) $\mathcal{X}$. Learning the hyperparameters $h$ of a GP and subsequent conditioning leads to a stochastic representation of a model function which can be used for decision-making, visualizations, and interpretations.

\vspace{2mm}
\noindent
%%%what's great about GPs, what's thew problem?
In comparison to neural networks, GPs scale better with the dimensionality of the problem and provide more exact function approximations \cite{manzhos2021optimization}. Additionally, GPs provide the highly-coveted Bayesian uncertainty quantification on top of such function approximations. While some neural-network-based methods can estimate errors, these are most often not the result of rigorous Bayesian uncertainty quantification. Even so, GPs come with one difficult-to-circumvent problem: Due to their unfavorable scaling of $O(N^3)$ in computation and $O(N^2)$ in storage \cite{williams2006gaussian}, the applicability of GPs has largely been limited to small and moderate dataset sizes ($N$), which prevents the method from being used in many fields where large datasets are common and is a major disadvantage compared to other ML methods, e.g., neural networks. Those fields include many machine learning applications, earth, environmental, climate and materials sciences, and engineering. The numerical complexity of GPs stems from the need to store and invert a typically-dense covariance matrix \cite{williams2006gaussian}. While the direct inversion can be replaced by iterative linear system solves, the speedup is rather modest for dense covariance matrices.

\vspace{2mm}
\noindent
Methods to alleviate the GP's scaling issues exist but are mostly based on approximations. These workarounds fall into a few broad categories:
\begin{itemize}
    \item A set of local GP experts: The dataset is divided into subsets, each of which serves as input into separate GPs, and the resulting posteriors are then combined \cite{cohen2020healing,gao2020generalized}.
    This can also be interpreted as one large GP with a sparse (block-diagonal) covariance matrix. It is common to divide the dataset by locality,
    leading to the name ``local GP experts''.
    \item Inducing-points methods: Instead of inducing a sparse covariance matrix by picking subsets of the dataset, inducing-point methods place new points inside the domain, inducing a favorable data structure that translates into sparsity. The function values at those points are calculated via standard interpolation techniques. Popular examples of this approach include KISS-GP \cite{wilson2015kernel}, the predictive process \cite{banerjee2008gaussian, finley2009improving}, and fixed-rank Kriging \cite{cressie2008fixed}. Generally, inducing-point methods are not agnostic to the kernel definition and therefore limit which kernels can be used. This limitation is a major drawback given that recent applications are increasingly using flexible non-stationary kernel functions (for instance \cite{remes2017non}), which are generally incompatible with inducing-point methods.
    \item Structure-exploiting methods: A special kind of inducing point method that places pseudo-points on grids, so that the covariance matrix has Toeplitz algebra, which leads to fast linear algebra needed to train and condition the GP.
    Again, the success of those methods is not agnostic to the kernel definition. 
    \item Vecchia approximations: Instead of calculating the full conditional probability density function of a GP prior, the Vecchia approximation \cite{vecchia1988estimation,katzfuss2021general} is used to pick a subset of the data to condition on. This method is also kernel-dependent and has largely been applied for stationary kernels.  
\end{itemize}
The statistics literature contains a variety of other related approaches; see \cite{heaton2019case} for a recent summary of both traditional and state-of-the-art approaches with a direct comparison of the methods on a common dataset. All of the existing methods introduced above have one thing in common: sparsity or exploitable structure in the covariance matrix is introduced by operating on the data points --- either by considering subsets of the full dataset or by utilizing representative pseudo-points that allow for a favorable structure (e.g. Toeplitz) in the covariance, and sparsity. This commonality leads to one major issue of all existing methods: they are approximations of exact GPs \cite{wang2019exact}, which leads to poor prediction performance for highly non-linear functions --- i.e. functions exhibiting large first and second-order derivatives with frequently changing signs. For high-fidelity approximations, the number of sub-selected data points or pseudo points must approach the size of the original dataset \cite{cohen2020healing},  which eliminates the methods' advantages. More fundamentally, the sparsity and structure of the covariance matrix should be dictated by the nature of the problem and the data, not by our computational constraints. This leads us to consider kernels that can take advantage of naturally occurring --- problem and data dictated --- sparsity.

\vspace{2mm}
\noindent
Instead of operating on the input points --- by selecting subsets or pseudo-points --- an alternative approach is to let the kernel find the most expressive and sparse structure of the covariance matrix. In principle, a very flexible kernel could discover --- not induce --- naturally occurring sparse structure in the covariance matrix without acting on the data points at all. In that case, there is no approximation taking place (compared to inducing-point, multiple-expert, and Vecchia methods) and no ad-hoc point selection is required.  Additionally, we shall see that there are no restrictions on the used problem-specific kernel functions as long as they are combined with our proposed sparsity-enabling, and therefore, sparsity-discovering kernels. An added advantage is that the kernel-discovered sparsity is entirely independent of spatial relationships of data points, meaning, very distant data points can be discovered to have high covariances while points in close proximity might be independent; there is no ad-hoc dependency of covariances on Euclidean point distance in $\mathcal{X}$ --- in contrast to local GP experts for instance.

\begin{figure}
\centering
     \begin{subfigure}[b]{0.8 \textwidth}
         \centering
         \includegraphics[width=\textwidth,trim={1cm 4cm 0cm 8cm},clip]{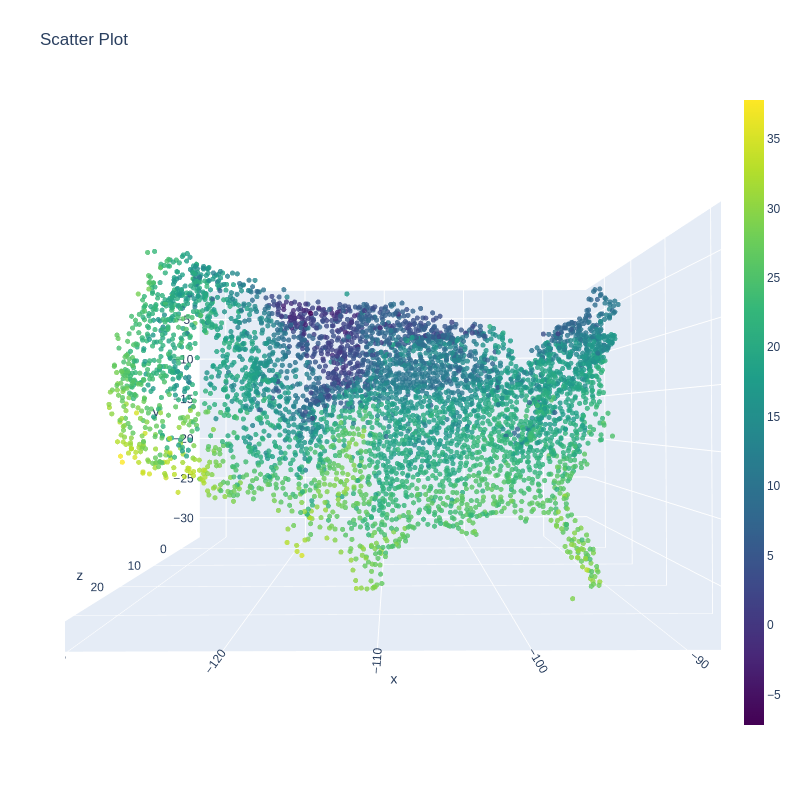}
         \caption{}
     \end{subfigure}

     \begin{subfigure}[b]{0.45 \textwidth}
         \centering
         \includegraphics[width=\textwidth,trim={1cm 1cm 0cm 3cm},clip]{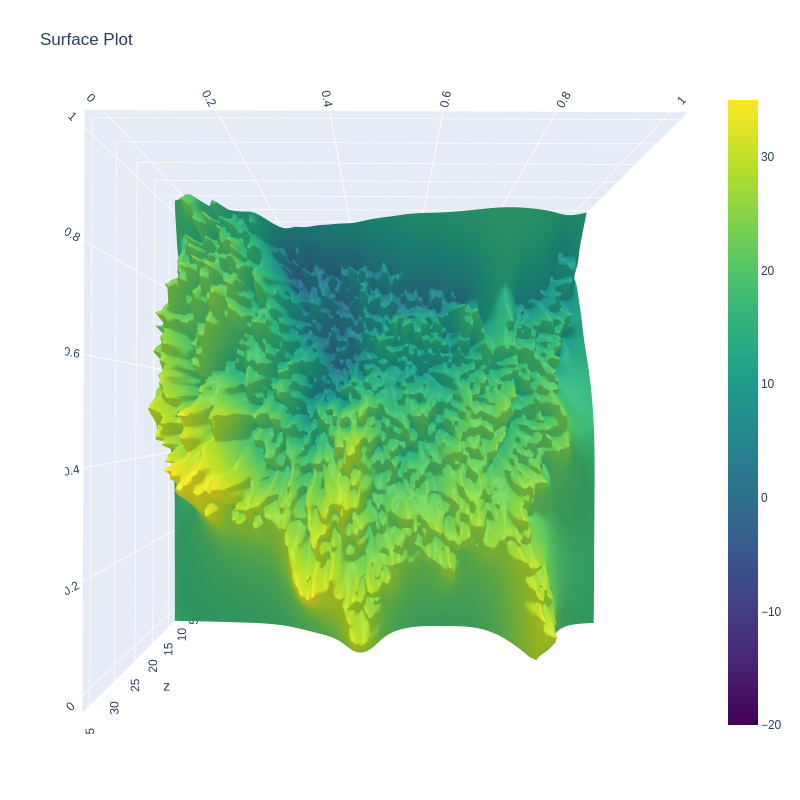}
         \caption{}
     \end{subfigure}
     \hfill
     \begin{subfigure}[b]{0.45 \textwidth}
         \centering
         \includegraphics[width=\textwidth,trim={1cm 1cm 0cm 3cm},clip]{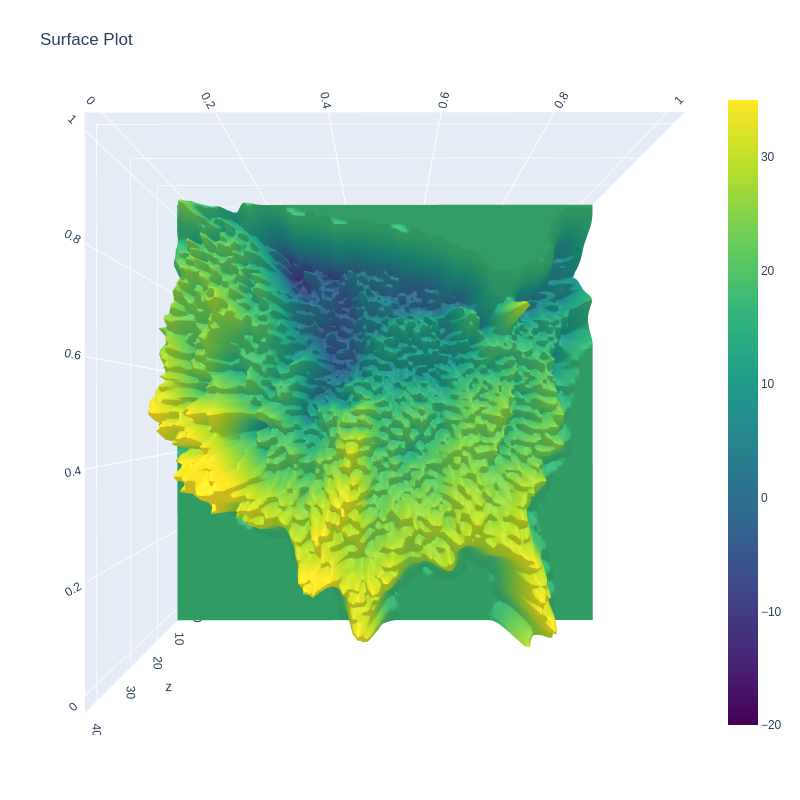}
         \caption{}
     \end{subfigure}
     
     \begin{subfigure}[b]{0.45 \textwidth}
         \centering
         \includegraphics[width=\textwidth,trim={1cm 1cm 0cm 3cm},clip]{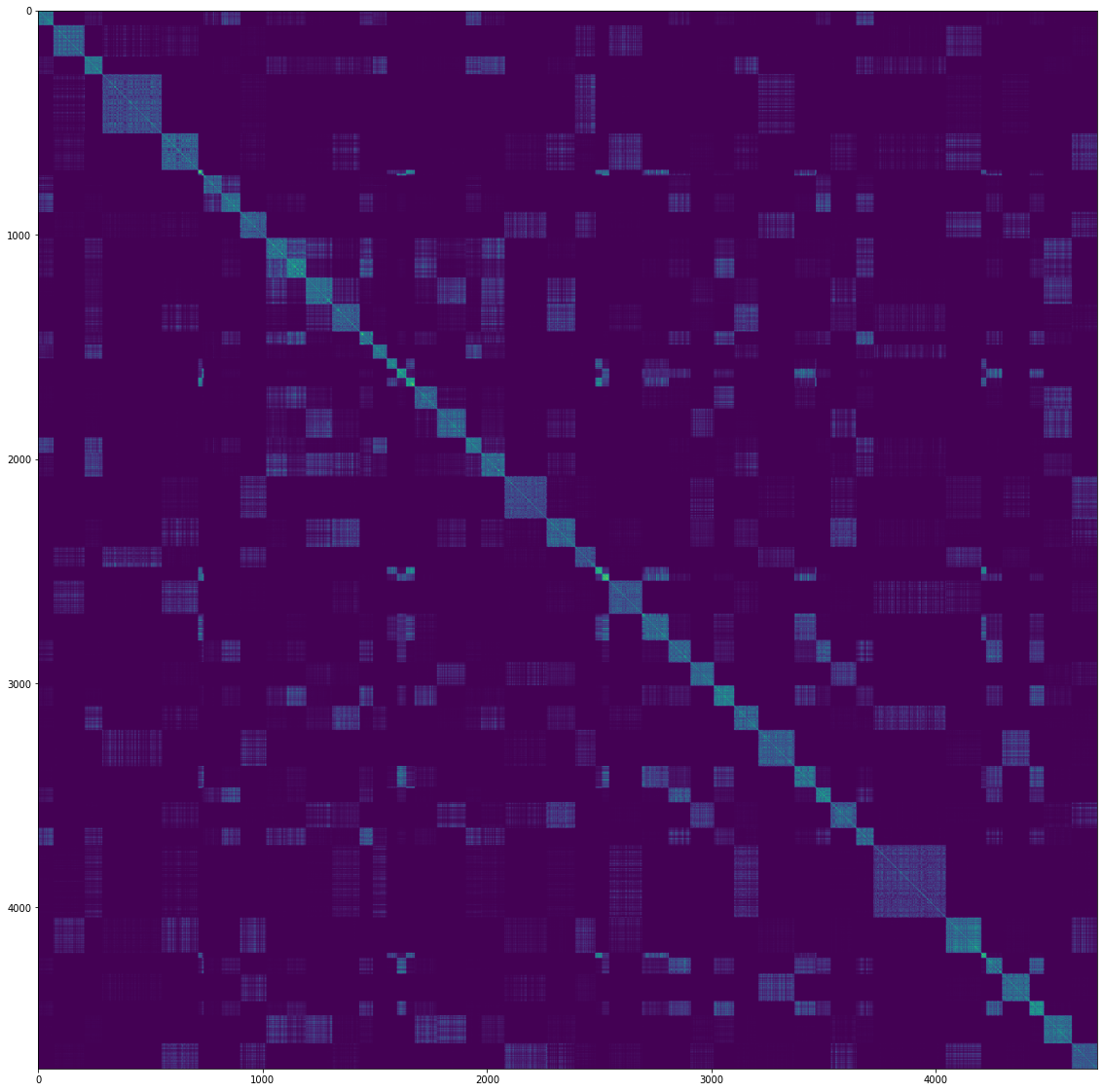}
         \caption{}
     \end{subfigure}
     \hfill
     \begin{subfigure}[b]{0.45 \textwidth}
         \centering
         \includegraphics[width=\textwidth,trim={1cm 1cm 0cm 3cm},clip]{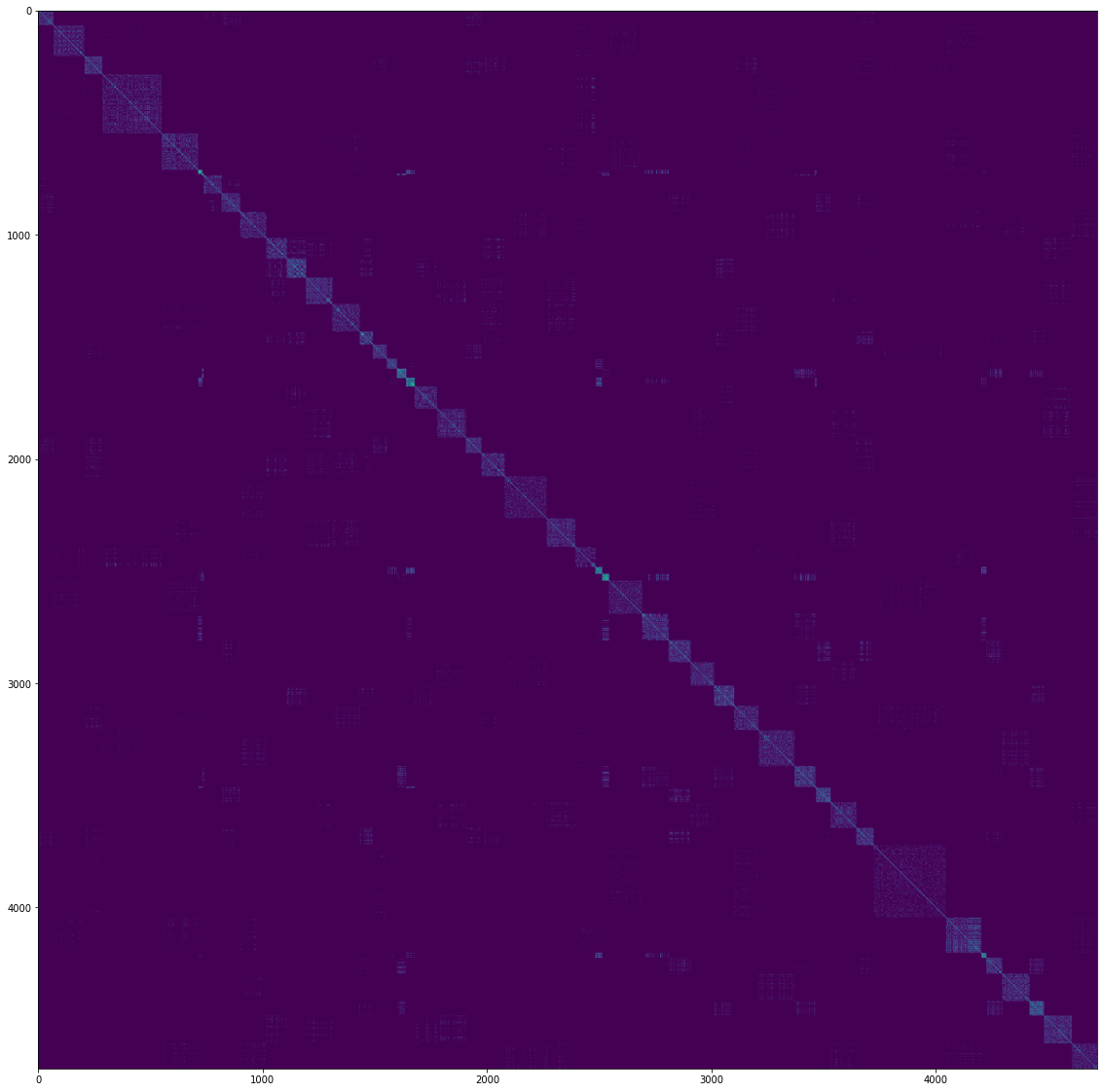}
         \caption{}
     \end{subfigure}
     %\put(-455,15){\rotatebox{90}{something}}
    \caption{\small Figure illustrating the premise of our proposed algorithm. (a) Shows the test data, measured daily maximum temperatures ($^\circ$C) from April 10th, 1990 across the United States. The dataset contains 4718 points. This problem size is still well within the capabilities of a standard GP, whose posterior mean is shown in (b). If we employ a flexible, non-stationary, and compactly-supported kernel we can learn through optimization of the marginal-log likelihood that only a few covariances are of essence for the prediction. Our sparse result is shown in (c). The corresponding covariance matrix only has 1.5\% of the non-zero entries of the full dense matrix. The sparsity in this problem is discovered, not induced, leading to an exact GP. This principle, in combination with HPC, for truly large covariance matrices, and constrained function optimization, enable GPs to be scaled to tens of millions of data points.}
    \label{fig:premise}
\end{figure}

\vspace{2mm}
\noindent
As we outline below, creating an exact GP that learns and utilizes naturally-occurring sparsity shall require three main building blocks: (1) ultra-flexible, compactly-supported kernel functions, specially customized to learn and encode zero-covariances, (2) a high-performing implementation that can compute sub-matrices of the covariance matrix in a distributed, parallel fashion, and (3) a constrained or augmented optimization routine to ensure the learned covariance matrix is sparse (or at least enforce a preference for sparsity). This last point is important for large problems to protect the computing system from over-utilization. In the extreme case, in which naturally-occurring sparsity is insufficient or non-existent, having a sparsity-inducing optimization routine would seamlessly result in an optimal approximate GP. The contributions of this paper can be summarized as follows. We show that, by combining tailored kernel designs, HPC implementation, and constrained optimization, exact GPs can be scaled to datasets of any size, under the assumption of naturally-occurring sparsity. The core idea, that allows such scaling, is a sparsity-discovering kernel design and an optimization that learns which data points are not correlated, independent of their respective locality in the input space $\mathcal X$. Sparse structure is not artificially ``induced'' as in all state-of-the-art methods; instead, we allow the GP to discover the natural sparsity in the dataset. This principle is visualized in Figure \ref{fig:premise}.

%%%%%%%%%%%%%%%%%%%%%%%%%%%%%%%%%%%%
\section{Method}
%%%%%%%%%%%%%%%%%%%%%%%%%%%%%%%%%%%%
\subsection{Basics}
A Gaussian Process (GP) is characterized by a Gaussian probability density function over function values $\mathbf{f}$
\begin{equation}
    p(\mathbf{f})=\frac{1}{\sqrt{(2\pi)^\mathrm{dim}|\mathbf{K}|}}
    \exp \left[ -\frac{1}{2}(\mathbf{f}-\mathbf{m})^T \mathbf{K}^{-1}(\mathbf{f}-\mathbf{m}) \right],
    \label{eq:priorGP}
\end{equation}
and a Gaussian likelihood
\begin{equation}
    p(\mathbf{y}|\mathbf{f})=\frac{1}{\sqrt{(2\pi)^\mathrm{dim}|\mathbf{V}|}}
    \exp \left[ -\frac{1}{2}(\mathbf{y}-\mathbf{f})^T \mathbf{V}^{-1} (\mathbf{y}-\mathbf{f}) \right],
    \label{eq:likelihoodGP}
\end{equation}
where $\mathbf{V}$ is the observation-noise matrix, which is most often diagonal, $\mathbf{K}$ is the covariance matrix defined by the kernel function $\mathcal{K}_{i,j}~=~k(\mathbf{x}_i,\mathbf{x}_j)$, and $\mathbf{m}$ is the prior-mean vector. Training the GP
is done by maximizing the marginal log-likelihood (ignoring an additive constant)
\begin{equation}
    \ln(L(h))=-\frac{1}{2} (\mathbf{y}-\mathbf{m}(h))^T \mathbf{K}(h)^{-1} (\mathbf{y}-\mathbf{m}(h)) - \frac{1}{2} \ln(|\mathbf{K}(h)|)
    \label{eqn:log_likelihood}
\end{equation}
with respect to the hyperparameters $h$. After the hyperparameters are found, the posterior is defined as
\begin{align}\label{eq:pred_distr}
    p(\mathbf{f}_0|\mathbf{y})~&=~\int_{\mathbb{R^N}} 
    p(\mathbf{f}_0|\mathbf{f},\mathbf{y})~p(\mathbf{f}|\mathbf{y})~d\mathbf{f} \nonumber \\
    &~\propto \mathcal{N}(\mathbf{m}_0 +\pmb{\kappa}^T~
    (\mathbf{K}+\mathbf{V})^{-1}~(\mathbf{y}-\mathbf{m}_0 ), \pmb{\mathcal{K}} -
    \pmb{\kappa}^T~(\mathbf{K}+\mathbf{V})^{-1}~\pmb{\kappa}),
\end{align}
where $\pmb{\kappa}=k(\mathbf{x}_0,\mathbf{x}_j)$, and $\pmb{\mathcal{K}}=k(\mathbf{x}_0,\mathbf{x}_0)$.
This basic framework can be extended by ever-more flexible mean, noise and kernel functions.
Our proposed method is entirely agnostic and even symbiotic to those extensions and we will therefore omit the dependencies thereof.

\vspace{2mm}
\noindent
The bottleneck of training GPs, that is, optimizing \eqref{eqn:log_likelihood} with respect to $h$, is the $O(N^3)$ numerical complexity of calculating $\mathbf{K}(h)^{-1}\mathbf{y}$ --- or equivalently solving a linear system --- and $\ln(|\mathbf{K}(h)|)$, and the storage of $\mathbf{K}$, which scales $O(N^2)$. However, if $\mathbf{K}$ is very sparse, both problems would be avoided. This is the goal of all approximate methods, which work by synthetically inducing this sparsity.
In contrast to approximate techniques, we propose to achieve sparsity purely by flexible kernel design, and not through approximations, leading to a sparse but exact GP.
The sparsity, in this case, is discovered, not induced. However, if the problem does not have natural sparsity, the constrained optimization described below used to optimize \eqref{eqn:log_likelihood} shall guarantee the minimal approximations needed to satisfy system-dictated minimum-sparsity constraints.

\vspace{2mm}
\noindent
Consider, as a simple example, the squared exponential kernel
\begin{equation}
    k(\mathbf{x}_1,\mathbf{x}_2) = \sigma_s^2~\exp(-0.5~||\mathbf{x}_1 - \mathbf{x}_2||^2/l^2),
\end{equation}
which is used in about 90\% of GP applications \cite{pilario2020review}; even if data points were naturally uncorrelated, the squared exponential kernel would not be able to learn this independence. This is true for all commonly used stationary kernels and most non-stationary kernels. Instead of formulating kernels that learn well which points are dependent, we propose to consider kernels that are tailored to be capable of learning independence. Such a \textbf{non-stationary, flexible, and compactly-supported kernel is the first building block} of the proposed framework. Even if such a kernel can be defined, the covariance matrix still has to be computed and stored which is time-consuming and often prohibitive due to storage requirements. \textbf{Distributed computing on HPC compute architecture --- as the second building block} --- can help by splitting up the computational and storage burden. The \textbf{third building block, augmented and constrained optimization} can guarantee that sparse solutions are given preference, or are even a requirement. 

%%%%spelling and grammar checked up to here 11 AM PST, April 26th

%%%%%%%%%%%%%%%%%%%%%%%%%%%%%%%%%%%%%%%%%%%%%%%%%%%%%%%%%%%%%%%%%%%%%%%%%%%%%%%%%%%%%%%%%%%%%
\subsection{Building Block 1: Non-Stationary, Ultra-Flexible and Compactly-Supported Kernel Functions}
For natural sparsity to be discovered, a kernel function $k(\mathbf{x}_1, \mathbf{x}_2)$ should be designed such that it can flexibly encode correlations between data points, including instances where no correlations exist \cite{melkumyan2009sparse}. The kernel has to meet three requirements:
\begin{enumerate}
    \item Compact Support: This is the most obvious necessary property. Since we are attempting to discover zero covariances, the kernel has to be compactly supported.
    \item Non-Stationarity: Compactly-supported kernels have been used before, but mostly in the stationary case. However, in the stationary case, sparsity is only taken advantage of in an entirely local way --- i.e. only if a point happens to be far away from all other points can the covariance be zero. Such kernels are not able to learn more complicated distance-unrelated sparsity-exploiting dependencies. 
    \item Flexibility: To pick up on sparsity across geometries and distances, a kernel has to be flexible to recognize that neighboring points may be correlated and some points in the distance are not, and vise versa.
\end{enumerate}
Combining compact support, non-stationarity and flexibility yields kernels that are tailored to learn existing and non-existing covariances. Below we examine a few examples to solidify this idea. The kernel
\begin{equation}\label{eq:sparsek1}
    k_s(\mathbf{x}_1,\mathbf{x}_2) =\tilde{k}(\mathbf{x}_1,\mathbf{x}_2)~f(\mathbf{x}_1)f(\mathbf{x}_2),
\end{equation}
is a rather-well-known example of a non-stationary kernel. The subscript ``s'' stands for ``sparsity'', since this is the kernel that allows the discovery of sparsity. The kernel $\tilde{k}$ is assumed to be compactly-supported and stationary (for instance \cite{melkumyan2009sparse}); the non-stationarity is produced by the term $f(\mathbf{x}_1)f(\mathbf{x}_2)$. The flexibility of this kernel depends entirely on the parameterization of $f$. In the most flexible case, $f$ could be a sum of Kronecker-$\delta$ functions centered at a subset the data points $\hat{\mathcal D} \subseteq \left\{\mathbf x_i\right\}_{i=1}^N$, i.e.,
$$f(\mathbf x) = \sum_{\mathbf{x}_i \in \hat{\mathcal D}} h_i \delta(\mathbf{x}, \mathbf{x}_i),$$
where the $|\hat{\mathcal D}|$ binary coefficients $h_i \in \left\{0, 1\right\}$ are hyperparameters that may be optimized during training. If we allowed $\hat{\mathcal D}$ to include \emph{all} data points, we would obtain a GP that has learned which such points can safely be ignored $(h_i = 0)$ without impacting the marginal log-likelihood. 

\vspace{2mm}
\noindent
The kernel \eqref{eq:sparsek1} is very flexible but has two issues. First, it explicitly depends on potentially millions of binary hyperparameters. Second, it is unable to encode varying covariances between data points; points are either `turned on' or `off'. An even more flexible kernel, that can in fact turn on and off selected covariances instead of just points, can be defined as
\begin{equation}
    k_s(\mathbf{x}_1,\mathbf{x}_2) =\tilde{k}(\mathbf{x}_1,\mathbf{x}_2)~(f(\mathbf{x}_1)f(\mathbf{x}_2) + g(\mathbf{x}_1)g(\mathbf{x}_2)),
    \label{eq:sparsek2}
\end{equation}
where
\begin{align}
    f(\mathbf{x}) &= \sum_{\mathbf x_i \in \hat{\mathcal D}} h^f_i \delta(\mathbf x, \mathbf x_i) \\
    g(\mathbf{x}) &= \sum_{\mathbf x_i \in \hat{\mathcal D}} h^g_i \delta(\mathbf x, \mathbf x_i)
    \label{eq:sparsek2b}
\end{align}
where the $h^f_i$ and $h^g_i \in \left\{0, 1\right\}$ or $\in~[0,\infty]$. This kernel can effectively discover that certain covariances (perhaps most) are zero. 
If we include all data points in  $\mathcal D$, then this kernel has $2N$ hyperparameters to optimize, which can be an overwhelming optimization if $N$ is large.

\begin{figure}
    \centering
     \begin{subfigure}[b]{0.45 \textwidth}
         \centering
         \includegraphics[width=\textwidth,trim={3cm 2cm 4cm 6cm},clip]{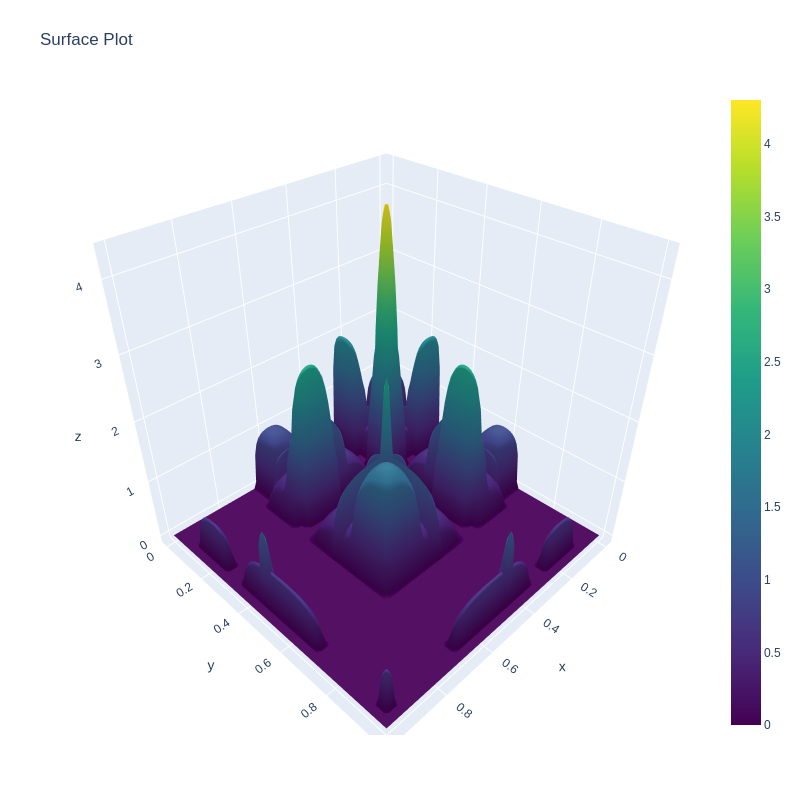}
         \caption{}
     \end{subfigure}
     \hfill
     \begin{subfigure}[b]{0.45 \textwidth}
         \centering
         \includegraphics[width=\textwidth,trim={3cm 2cm 4cm 6cm},clip]{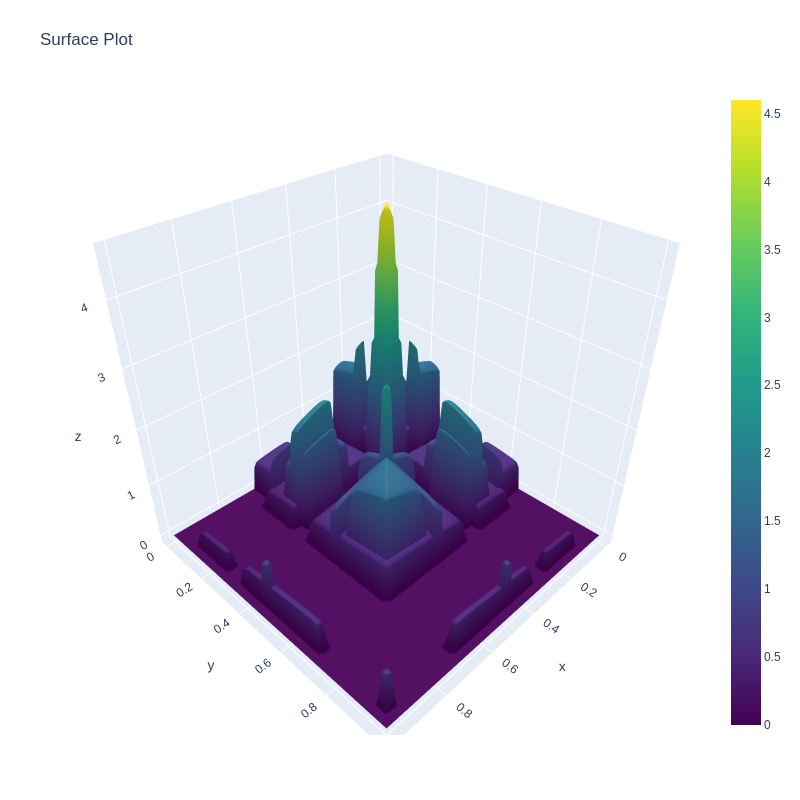}
         \caption{}
     \end{subfigure}
    \caption{\small Figure showing a very flexible, non-stationary and compactly-supported kernel function $k(x,y)$ (a, \eqref{eq:bump_kernel}) , and a ``sparsified'' squared exponential core kernel $k_c$ (b). Only for a one-dimensional input domain, we can visualize the kernel as a function over $\mathcal{X}\times\mathcal{X} \subset \mathbb{R}\times\mathbb{R}$. The kernel uses a set of compactly-supported bump functions to naturally discover sparsity through optimization of the bump functions' positions, heights, radii, and shapes. Since any multiplications of kernels is a valid kernel, our sparsity-discovering kernel $k_s$ can be combined with any kernel; therefore, compared to most approximate method, it does not limit the user's ability to design and employ arbitrary kernel functions.
    Panel (b) shows that concept; where the kernel $k_s$ has support, the covariance function becomes the squared-exponential kernel.  
    }
    \label{fig:kernel}
\end{figure}

\vspace{2mm}
\noindent
To alleviate the challenge of a large number of hyperparameters, we can trade some of the flexibility and therefore sparsity for a parameterization with fewer hyperparameters. For this purpose we propose the kernel function
\begin{equation}
     k_s(\mathbf{x}_1,\mathbf{x}_2) =\tilde{k}(\mathbf{x}_1,\mathbf{x}_2)~ \sum_i^{n_1} f_i(\mathbf{x}_1)f_i(\mathbf{x}_2),
     \label{eq:bump_kernel}
\end{equation}
where 
\begin{equation}
    f_i(\mathbf x) = \sum_{j=1}^{n_2} a_{ij} \exp\left[\frac{-\beta_{ij}}{1-\frac{||\mathbf{x}-\mathbf{x}_0^{ij}||^2_{2}}{r_{ij}^2}} + \beta_{ij} \right] \chi(r_{ij} < ||\mathbf{x}-\mathbf{x}_0^{ij}||_{2}).
    \label{eq:bump}
\end{equation}
Equation \eqref{eq:bump} is a sum of, so-called, bump functions, where $\chi(r_{ij} < ||\mathbf{x}-\mathbf{x}_0^{ij}||_{2})$ is the indicator function which is 1 if $r_{ij} < ||\mathbf{x}-\mathbf{x}_0^{ij}||_{2}$ and 0 otherwise, $\mathbf{x}_0^{ij}$ are the bump function locations, $r_{ij}$ are the radii, and $\beta_{ij}$ are shape parameters. Bump functions are $\in~C^{\infty}$ and compactly supported; precisely the properties we need to create sparsity-discovering kernel functions. The kernel function \eqref{eq:bump_kernel} allows us to seamlessly choose between flexibility, which directly impacts the ability to discover sparsity, and the number of hyperparameters (compare \eqref{eq:bump_kernel} with \eqref{eq:sparsek2} for $n_1=2$ and $n_2=1$). Figure \ref{fig:kernel} shows a visualization of this kernel. For our test, we will combine the above kernel with a compactly-supported stationary kernel given by
\begin{equation}\label{eq:kstat}
    \tilde{k}(\mathbf{x}_1,\mathbf{x}_2) = 
    \begin{cases}
    \frac{\sqrt{2}}{3 \sqrt{\pi}} 
    \Big( 
    (3 (\frac{d}{r})^2 \log{\left( \frac{  \frac{d}{r}}{1+\sqrt{1 - (\frac{d}{r})^2}}\right)}+\
    (2 (\frac{d}{r})^2+1) \sqrt{1-(\frac{d}{r})^2} 
    \Big) ~ \text{if }   d<r, \\
    0 \text{ else} 
    \end{cases}
\end{equation}
where $d=||\mathbf{x}_1-\mathbf{x}_2||_{2}$, and $r$ is the radius of support. Kernel function \eqref{eq:kstat} is a rather-well-known compactly-supported stationary kernel. Since kernels can be multiplied, we can combine our sparsity-discovering kernel $k_s$ \eqref{eq:bump_kernel} with any other kernel ($k_c~\cdot~k_s$), leading to no restrictions on the core kernel $k_c$. $k_s$ can be normalized and shaped in order to equal one within its support and zero otherwise, which can then be understood as a mask that leaves the core kernel $k_c$ untouched in areas of support. The kernel $k_s$ also gives us the opportunity to estimate the sparsity of the covariance matrix. In the limit of infinitely many, uniformly distributed data points, the sparsity $s$ of the covariance matrix is bounded from above as
%spelling and grammar checked April 26, 12:18
\begin{equation}
    s=\frac{\text{number of non-zero covariances}}{N^2} \leq \frac{\int_{S_k}~d\mathbf{x}d\mathbf{x}}{\int_{\mathcal{X}\times \mathcal{X}}~d\mathbf{x}d\mathbf{x}},
    \label{eq:sparsity_estimate}
\end{equation}
which we can use to formulate objective functions that allow us to give preference to sparse solutions, or to formulate sparsity constraints; note that a small $s$ here means high sparsity. $S_k$ in \eqref{eq:sparsity_estimate} is the set of support of the kernel, i.e., $S_k \subset \mathcal{X}\times \mathcal{X}$, which, in our case is the Cartesian product of two balls $\mathcal{B}\subset\mathcal{X}$ \footnote{The Volume of the Cartesian-product set of two balls embedded in $\mathbb{R}^n$, $\mathcal{B}_1 \times \mathcal{B}_1$ is the product of their respective volumes.}. 
Therefore, for kernel \eqref{eq:bump_kernel}, the sparsity can be bounded from above --- assuming entirely disjoint support since any overlap increases sparsity (lowers $s$) --- so that
\begin{equation}
    \sup\{\int_{S_k}~d\mathbf{x}d\mathbf{x}\}=\sum_i^{n_1} \sum_j^{n_2} \sum_k^{n_2} Vol_s(dim,r_{ij})Vol_s(dim,r_{ik}),
\end{equation}
where $Vol_s(dim,r)$ is the volume of a $dim$-dimensional sphere with radius $r$, defined as
\begin{equation}
    Vol_s(dim,r)=\frac{\pi^{dim/2}}{\Gamma(\frac{dim}{2} + 1)} r^{dim},
\end{equation}
where $\Gamma$ is the gamma function, and $dim$ is the dimensionality of $\mathcal{X}$.
As $\beta\rightarrow 0$ in equation \eqref{eq:bump}, the kernel's affect on $k_c$ in regions of support vanishes (see \ref{fig:kernel} for an example). $\beta$ can therefore be seen as a shape parameter. As $\beta\rightarrow \infty$, the bump functions become delta functions and kernel \eqref{eq:sparsek2} is obtained.

%%%%%%%%%%%%%%%%%%%%%%%%%%%%%%%%%%%%%%%%%%%%%%%%%%%%%%%%%%%%%%%%%%%%%%%%%%%%%%%%%%%%%%%%%%%%%
\subsection{Building Block 2: High Performance Computing to Take Advantage of Sparse Kernels}
While flexible non-stationary and compactly-supported kernels are the core building block of our algorithm for extreme-scale GPs, the covariance matrix has to be computed in a dense format first to take full advantage of multi-threading, however, this could violate RAM restrictions for large datasets; computing the covariance matrix in a sparse format in place would be prohibitively inefficient. To avoid slow computations or going beyond the RAM limit, we define a ``host'' covariance matrix (on one host machine) as sparse in the first place, compute dense sub-matrices in a distributed way and cast them into a sparse format, and only communicate sparse sub-matrices back to the host machine, where they are inserted into the host covariance matrix. Through this strategy, we address RAM limitations by distributing the covariance matrix across many computing resources --- and could even exploit out-of-core methodologies such as utilizing disk storage if needed. Additionally, the computation time is sped up by leveraging heterogeneous architectures such as GPUs, efficient at data-parallel operations, and threading-task-parallel CPU operations. The combination of distributing memory and exploiting parallelism across cores allows our algorithm to operate on datasets of practically-unlimited size --- given enough distributed workers and sufficient natural sparsity. The procedure is illustrated in Figure \ref{fig:comp} and shown in pseudo-code \ref{alg:comp}.

%spell and grammar checked 12:40, April 26th
\vspace{2mm}
\noindent
We split up the dataset of length $|\mathcal{D}|$ into batches of size $b$. 
For large $|\mathcal{D}|$, the only way the covariance matrix can be computed is by distributing the computational burden by dividing the host covariance matrix into
block sub-matrices, each representative of a unique data-batch pair (see Figure \ref{fig:comp}). The batch pairs are transmitted to different workers (often a few per node) via DASK; we denote the number of parallel-executed tasks by $n$ (one task per worker). In each task, the exact batch-covariance is computed. Because of the specifically-designed kernel, many elements of each sub-matrix will be zero. That way, theoretically, any-size covariance matrix can be computed and stored in a distributed way. As the sub-matrices are transferred back, they will be translated into a sparse representation and injected into the sparse host covariance matrix on the host machine. While this matrix is $|\mathcal{D}| \times |\mathcal{D}|$ in size, its sparsity avoids problems with storing or computations. The computation of a batch of the covariance matrix can be accelerated by taking advantage of the many parallel threads a GPU or CPU has to offer. Future work will compare the compute performance of different implementations and architectures. 

\vspace{2mm}
\noindent
The proposed algorithm, given more resources, is able to compute solutions faster, exhibiting the strong scaling properties inherent in the design (see Figure \ref{fig:scaling}). Furthermore, as the problem size increases, the algorithm matches the set of resources also highlighting weak scaling. In summary, our formulation speeds up computation, reduces memory burden, and provides an ability to exploit heterogeneous architectures (CPUs/GPUs/TPUs), providing future compatibility of the proposed framework since future architectures can be leveraged.

%spelling and grammar checked 13:37, April 26th
\vspace{2mm}
\noindent
The theoretical computing time of the covariance matrix can be calculated as
\begin{equation}
    T_c=\frac{|\mathcal{D}|}{2nb}(\frac{|\mathcal{D}|}{b}+1)~t_b,
    \label{eq:scaling1}
\end{equation}
where $t_b$ is the compute time for one sub-matrix, whose scaling depends on the exact implementation, and availability and number of parallel CPU or GPU threads. Equation \eqref{eq:scaling1} suggests that, as the number of tasks $n$, the number of parallel workers, approaches $\frac{|\mathcal{D}|}{b}$, the scaling becomes linear in $|\mathcal{D}|$, i.e. complexity $O(|\mathcal{D}|)$. By extension, as the number of workers approaches the total batch number, the scaling becomes constant. The linear-system solution can be accomplished by the conjugate gradient method which has numerical complexity $O(m\sqrt{k})$, where $m$ is the number of non-zero entries in the covariance matrix and $k$ is the condition number. The log-determinant computation can be done via Cholesky-factorization whose scaling depends on the exact structure of the matrix. Furthermore, since for most intents and purposes $\frac{|\mathcal{D}|}{b}>>1$, we can approximate
\begin{equation}
    T_c \approx \frac{|\mathcal{D}|^2 t_b}{2nb^2},
    \label{eq:scaling2}
\end{equation}
which can help estimate the optimal batch size given a particular architecture. For sequential computations, $t_b$ scales $O(b^2)$ and the batch size drops out of the equation. For the other extreme, perfect parallelization, $t_b$ scales $O(b)$ and we therefore want to maximize the batch size up to the point where the linear scaling stops. That number depends on the particular architecture.

\begin{figure}
    \centering
    \includegraphics[width = \linewidth]{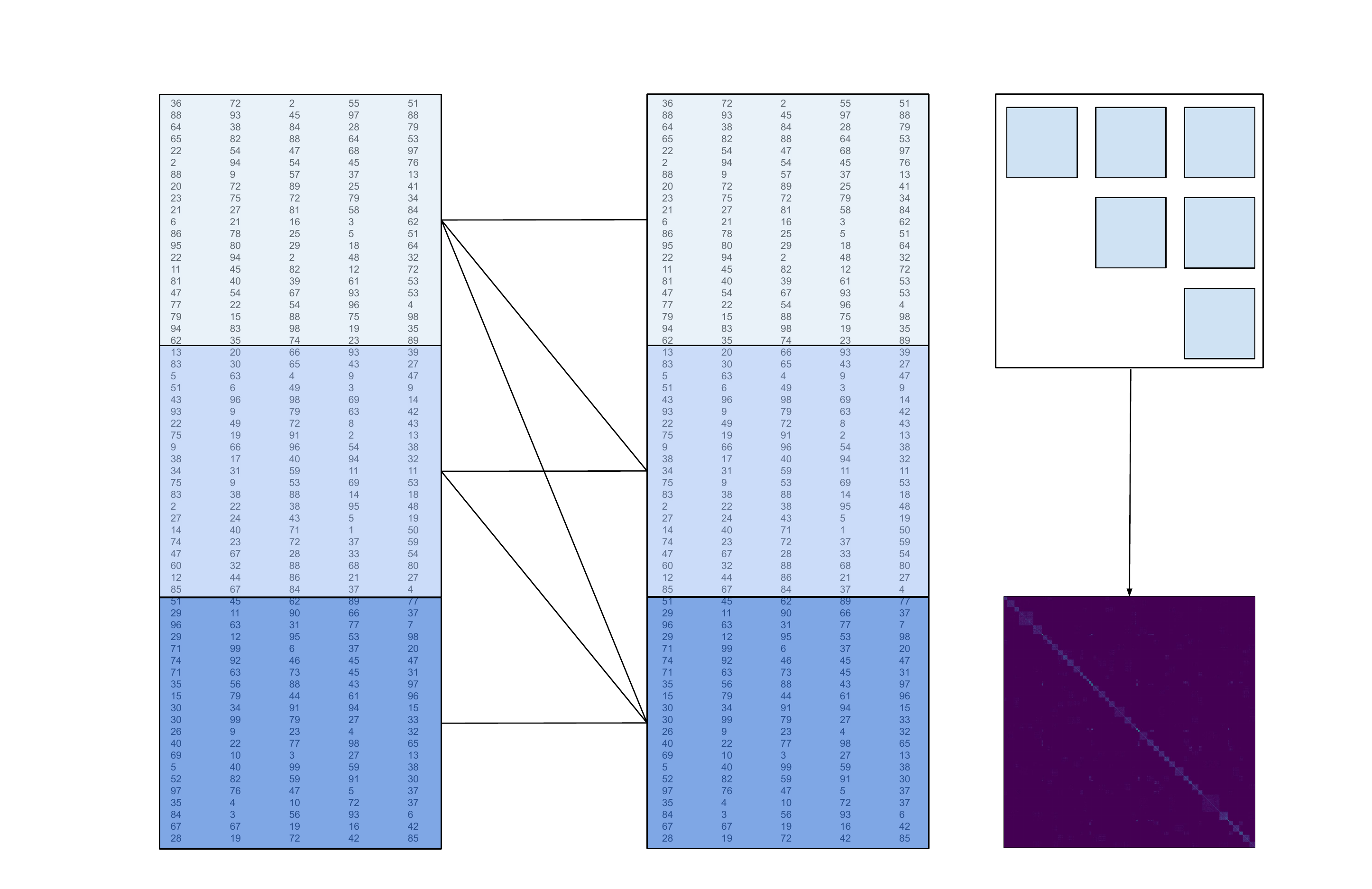}
    \put(-350,260){Dataset}
    \put(-200,260){Dataset}
    \put(-288,240){Compute Node}
    \put(-260,210){1}
    \put(-260,180){2}
    \put(-260,140){3}
    \put(-260,110){4}
    \put(-260,80){5}
    \put(-260,40){6}
    \put(-400,190){\rotatebox{90}{Batch A}}
    \put(-400,110){\rotatebox{90}{Batch B}}
    \put(-400,30){\rotatebox{90}{Batch C}}
    \put(-111,230){A,A}
    \put(-83,230){A,B}
    \put(-55,230){A,C}
    \put(-83,200){B,B}
    \put(-55,200){B,C}
    \put(-55,172){C,C}
    \put(-95,70){\textcolor{white}{\small Full Sparse}}
    \put(-95,60){\textcolor{white}{\small Covariance}}
    \caption{\small Figure illustrating the computational building block of the proposed algorithm. The dataset is divided into batches. Pairs of batches are sent to the compute nodes where the associated sub-matrices of the covariance matrix are calculated using the presented sparse kernels (equation \eqref{eq:bump_kernel}). The sub-matrices are cast into a sparse format on the compute nodes before being sent back to the host. There, they get assembled to obtain the full sparse master covariance matrix. All subsequent mathematical operation needed for a GP, namely calculating the log-determinant and solving a linear system, are performed efficiently on the sparse covariance matrix.}
    \label{fig:comp}
\end{figure}
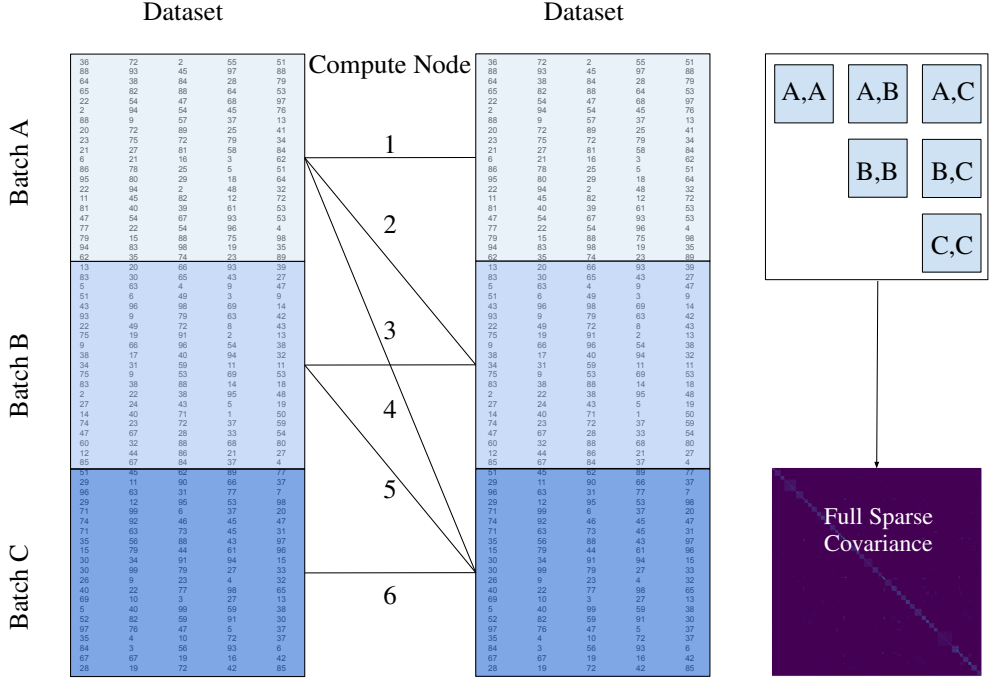
\begin{figure}
\centering
     \begin{subfigure}[b]{0.8 \textwidth}
         \centering
         \includegraphics[width=\textwidth,trim={0cm 0cm 2cm 3cm},clip]{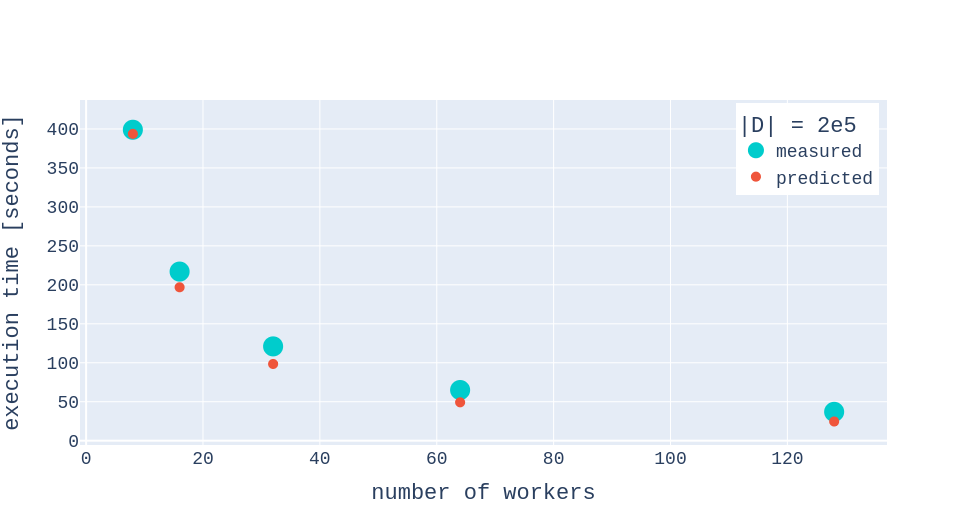}
         \caption{}
     \end{subfigure}
     
     \begin{subfigure}[b]{0.8 \textwidth}
         \centering
         \includegraphics[width=\textwidth,trim={0cm 0cm 2cm 3cm},clip]{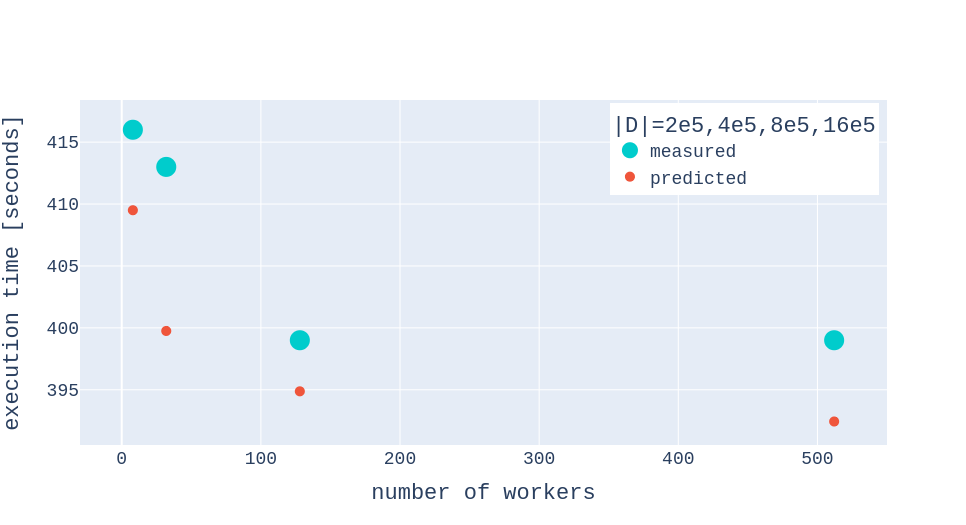}
         \caption{}
     \end{subfigure}
    \caption{\small Figure illustrating theoretical and measured strong and weak scaling of the distributed covariance computation. (a) Computation time of a problem of fixed size as a function of the number of workers. (b) Computation time as a function of the number of workers while the problem is increased (see label). The figures suggest that there is a strong case to be made for scalability of exact Gaussian processes.}
    \label{fig:scaling}
\end{figure}

\begin{algorithm}
\caption{Distributed Covariance Computation}\label{alg:comp}
\begin{algorithmic}[1]
\Procedure{Compute Covariance}{}
\State SparseCovariance = sparse matrix(dataset size,dataset size)
\State tasks = []
\For {i = 0:number of batches}{ in parallel}
\For {j = i:number of batches}{ in parallel}
\State select worker
\State tasks.append(dask.distributed.client.submit(kernel,batches,batch\_coordinates,hyperparameters))
\State SparseCovariance, tasks $=$ collect\_submatrices(SparseCovariance,tasks)
\EndFor
\EndFor
\State client.gather(tasks)
\EndProcedure
\State
\Procedure{collect\_submatrices}{SparseCovariance,tasks}
\State new\_tasks = []
\For {task in tasks}
\If {task is finished}
\State submatrix = task.result() \#only sparse matrices are communicated
\State SparseCovariance $=$ insert(submatrix)  \#insert the sub-matrix into host covariance matrix
\Else
\State new\_tasks.append(task)
\EndIf
\EndFor
\State return SparseCovariance, new\_tasks
\EndProcedure
\end{algorithmic}
\end{algorithm}
%grammar and spelling checked, April 26, 13:49 
%%%%%%%%%%%%%%%%%%%%%%%%%%%%%%%%%%%%%%%%%%%%%%%%%%%%%%%%%%%%%%%%%%%%%%%%%%%%%%%%%%%%%%%%%%%%%
\subsection{Building Block 3: Augmented and Constrained Optimization}
The proposed kernel definition \eqref{eq:bump_kernel} can be paired with constrained
\begin{align}
    &\argmax_{h} \ln(L) \nonumber \\
    &\text{subject to }s~<~\text{sparsity requirement}
    \label{eq:constrained}
\end{align}
or augmented optimization
\begin{equation}
    \argmax_{h,s} \Big( \ln(L) + (1-s)\ln(L) \Big),
    \label{eq:augmented}
\end{equation}
where $s$ is the estimated sparsity (equation \ref{eq:sparsity_estimate}).
The constraint means that this formulation will only be exact up until the RAM restriction is hit, then the GP will turn itself into an approximate GP, but without the need for the user to make decisions on which points are being considered.
The augmented (or biased) optimization will always prefer sparse covariances matrices. However, caution has to be exercised to ensure the optimization is not dominated by the need for sparsity. The formulation in equation \eqref{eq:augmented} gives priority to the likelihood, since $s\in[0,1]$, which means the objective function is bounded by $[\ln(L),2\ln(L)]$.

\subsection{A Note on Solving Linear Systems, Log-Determinants, and Optimization Strategies}
After computing the sparse covariance matrix in a distributed fashion, all that is left to do to enable GP training is to optimize the marginal log-likelihood. For this, we need to solve
\begin{equation}
    \mathbf{K}\mathbf{x}=\mathbf{y}
\end{equation}
and compute
\begin{equation}
    \log(|\mathbf{K}|).
\end{equation}
A common approach in the dense and the sparse case is to use Cholesky or LU factorization. Given the factorization, both the linear-system solution and the log-determinant computation is trivial. However, even for a sparse input matrix, both Cholesky and LU might have large memory requirements depending on fill-in and pivoting options. In addition, for those decomposition methods to be successful, the matrices have to be extremely sparse with only a handful of non-diagonal non-zero entries; a level of sparsity we might not be able to guarantee for matrices originating from a GP. 
In our experience, it is better to use iterative methods (e.g. conjugate gradients) to solve the linear system. This leaves us with the problem of estimating the log-determinant accurately.
For this work, we have employed random linear algebra (RLA). More specifically, we have implemented the method presented in \cite{boutsidis2017randomized}. Since we are training the GP via Markov-Chain Monte-Carlo the random noise induced by RLA won't affect the training. As we move to more deterministic optimizers, especially derivative-based optimizers, this discussion will have to be revisited.  
%grammar and spelling checked, April 27, 07:38
%%%%%%%%%%%%%%%%%%%%%%%%%%%%%%%%%%%%
\section{A Climate Example with over 5 Million Data Points}
%%%%%%%%%%%%%%%%%%%%%%%%%%%%%%%%%%%%
We demonstrate the proposed methodology on a temporal extension of the dataset shown in Figure \ref{fig:premise}. The dataset contains daily maximum temperatures from 1990 to 2019 from circa 7500 gauge-based weather stations across the continental United States \citep{Menne2012,ghcnd_data}; after accounting for missing daily measurements, these stations yield over 51.6 million data points across this approximately 10000-day (30-year) period. 
Due to computing constraints, while writing this paper, we randomly extracted a dataset of 5165718 points to use for our example. For reproducibility purposes, the dataset can be found online at \\
\emph{ftp://ftp.ncdc.noaa.gov/pub/data/ghcn/daily/}. For our tests, we used the \emph{gp2Scale} library that is part of the \emph{fvGP} Python package, available from github (\emph{https://github.com/lbl-camera/fvGP}) and pypi (\emph{pip install fvgp}).

\vspace{2mm}
\noindent
A Gaussian process is needed to analyze these data for a variety of reasons. First, while in situ measurements of daily weather variables provide the most realistic data source for understanding historical climate, users of such data often require geospatially-interpolated datasets that account for irregular sampling density and provide a complete picture of how temperatures vary over space. Daily maximum temperatures furthermore exhibit strong spatial autocorrelations due to their driving physical mechanisms in the ocean and atmosphere, which GPs are particularly well-suited to model statistically. In the temporal domain, autocorrelations are also generally quite strong in daily temperatures due to, e.g., seasonality imposed by the solar cycle, and as such GPs are needed to appropriately impute missing measurements at the gauge locations.

\vspace{2mm}
\noindent
For this example, we defined the kernel as
\begin{equation}
    k(\mathbf{x}_1,\mathbf{x}_2)=\tilde{k} (f(\mathbf{x}_1)f(\mathbf{x}_2)+g(\mathbf{x}_1)g(\mathbf{x}_2)),
\end{equation}
where $\tilde{k}$ is defined in equation \eqref{eq:kstat}, and $f$ and $g$ are defined in equation \eqref{eq:bump} with $n_2 = 4$, giving rise to 42 hyperparameters.

\begin{figure}
\centering
     \begin{subfigure}[b]{0.45 \textwidth}
         \centering
         \includegraphics[width=\textwidth,trim={4cm 4cm 4cm 4cm},clip]{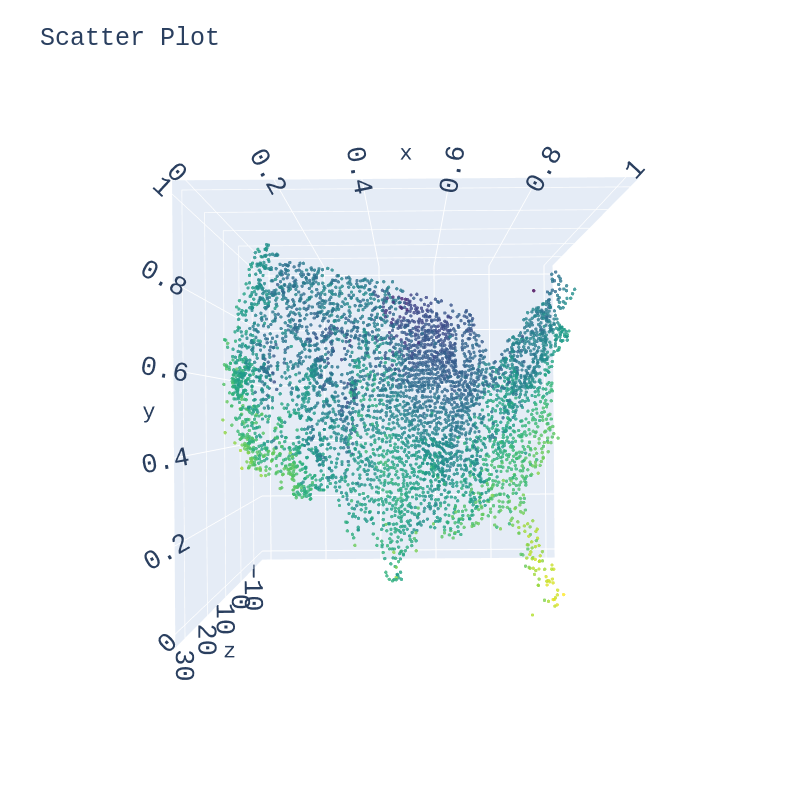}
         \caption{}
     \end{subfigure}
     \begin{subfigure}[b]{0.45 \textwidth}
         \centering
         \includegraphics[width=\textwidth,trim={2cm 0cm 0cm 2cm},clip]{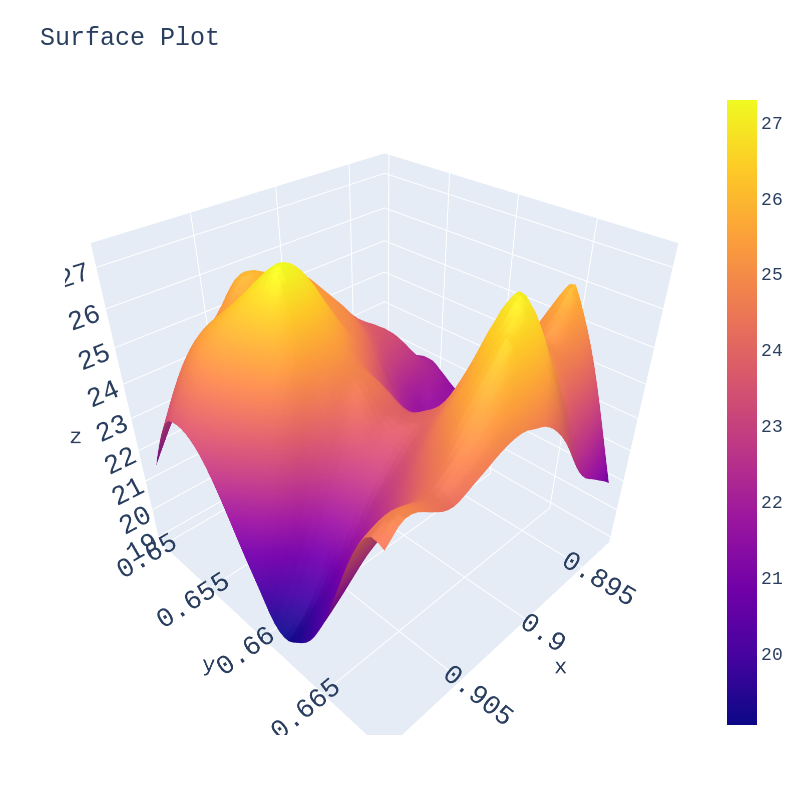}
         \caption{}
     \end{subfigure}
     
     \begin{subfigure}[b]{ \textwidth}
         \centering
         \includegraphics[width=\textwidth,trim={0cm 0cm 2cm 3cm},clip]{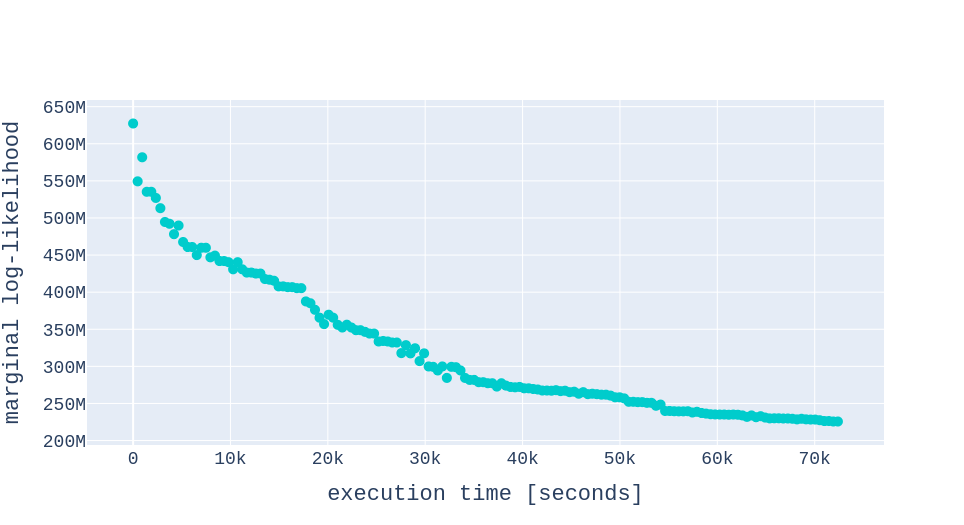}
         \caption{}
     \end{subfigure}
    \caption{\small The result of a Gaussian process trained on over 5 million data points. While this paper is best understood as a proof-of-concept, we want to ensure that we show the readers that the resulting model is reasonable by the end of our training (a,b). (a) The distributions of the climate stations with temperatures from the first day of the dataset (Jan 1st, 1990); the axes are normalized. (b) The GP interpolation over a subdomain in the north-east at a time slice in June 2004. The noise of the measurement was estimated ad-hoc, which explains the somewhat rough appearance of the posterior-mean function. We trained the GP via MCMC for 160 iteration. While this does not reach convergence it is enough to demonstrate the feasibility of such an extreme-scale GP. (c) Shows the marginal log-likelihood as a function of training time. The GP was trained in under 24 hours on 256 GPUs, opening the doors for much larger GPs.}
    \label{fig:exp}
\end{figure}

\vspace{2mm}
\noindent
To deliver a proof-of-concept of the proposed strategy, we are employing a Markov-Chain-Monte-Carlo training up to 160 function evaluations. Since the total compute time scales linearly
with the number of function evaluations, it is straightforward to estimate the compute time for many other training strategies. For this test, we chose two different architectures, namely Nersc's Cori Haswell Nodes \newline (\emph{https://www.nersc.gov/systems/cori/}) and Perlmutter's GPU nodes (Perlmutter Phase 1: \newline \emph{https://www.nersc.gov/systems/perlmutter/}). Due to challenges with allocating DASK workers on Cori, the result shown was calculated on 256 of Perlmutter's A100 Nvidia GPUs. Computing a batch of size 10000 can be accomplished in circa 0.6 seconds on each GPU node. See Figure \ref{fig:exp} for the visualization of the result. 

\vspace*{2mm}
\noindent
Due to early-access constraints, we split up this run into 4 separate runs, storing the hyperparameters and therefore the state of the training. 
Therefore, the total run time of 24 hours contains 4 initializations. Each iteration of the MCMC took circa 460 seconds, leading to a total estimated run time of 72384 seconds.

%%%%%%%%%%%%%%%%%%%%%%%%%%%%%%%%%%%%
\section{Summary, Discussion and Conclusion}
%%%%%%%%%%%%%%%%%%%%%%%%%%%%%%%%%%%%
In this paper, we have proposed a new methodology and algorithm for extreme-scale exact Gaussian processes (GPs) based on flexible, non-stationary and compactly-supported kernels, and distributed computing. Our method is not another approximate GP but is designed to discover --- not induce --- naturally-occurring sparsity and use it to alleviate challenges with numerical complexity in compute time and storage. It is our strong belief that this natural sparsity is very common in many modern datasets. The fundamental assumption in this work is that GPs often give rise to sparse covariance matrices naturally if given enough flexibility, through non-stationary kernel designs, to discover the sparsity. This can only be achieved for kernels that are very flexible, non-stationary, and compactly supported. For efficiency reasons, the covariance still has to be computed in a dense format first which is accomplished by distributing the workload over many CPU or GPU nodes. Constrained or augmented optimization is used to give sparse solutions priority or to constrain sparsity. These constraints only take effect when RAM or computing restrictions of the system are exceeded and would then turn the exact GP into an optimal sparse GP.

\vspace{2mm}
\noindent
This work is at a proof-of-concept stage; therefore, there are several challenges with the current form and these will be addressed in future work:
\begin{enumerate}
    \item The sparsity-discovering kernel for our examples was relatively simple. It has to be shown that much more flexible bump-function-based kernels can be formulated and their hyperparameters can be found robustly. However, there is a trade-off to consider; a more flexible kernel will lead to better detection of sparsity, but a more costly optimization of the hyperparameters. More hyperparameters also mean possible ill-posed optimization problems. 
    \item We have used MCMC for training, which means maximizing the marginal log-likelihood. The proposed method should be extended for gradient-based optimization of the hyperparameters.
    \item While our covariance matrix is computed in a distributed manner, the linear-system solutions and log-determinant computations are serialized even though most workers are idle and should be used for that task. However, the observed sparsity was found to be so substantial that the computations were not a bottleneck. 
\end{enumerate}
Despite those shortcomings, the method has shown its strength by training a Gaussian process on more than five million data points. This is, to our knowledge the largest exact GP ever trained. Given the strong and weak scaling shown in Figure \ref{fig:scaling} and predicted by equation \eqref{alg:comp}, we are confident that exact GPs on 100 million data points are currently possible. The code is available as part of the open-source python packages \emph{fvGP} and \emph{gpCAM}.

\paragraph{Acknowledgments}
The work was funded through the Center for Advanced
Mathematics for Energy Research Applications 
(CAMERA), which is jointly funded by the 
Advanced Scientific Computing Research (ASCR) 
and Basic Energy Sciences (BES) within the 
Department of Energy's Office of Science, under Contract No. DE-AC02-05CH11231. 
This work was further supported by the Regional and Global Model Analysis Program of the Office of Biological and Environmental Research in the Department of Energy Office of Science under contract number DE-AC02-05CH11231. This document was prepared as an account of work sponsored by the United States Government. While this document is believed to contain correct information, neither the United States Government nor any agency thereof, nor the Regents of the University of California, nor any of their employees, makes any warranty, express or implied, or assumes any legal responsibility for the accuracy, completeness, or usefulness of any information, apparatus, product, or process disclosed, or represents that its use would not infringe privately owned rights. Reference herein to any specific commercial product, process, or service by its trade name, trademark, manufacturer, or otherwise, does not necessarily constitute or imply its endorsement, recommendation, or favoring by the United States Government or any agency thereof, or the Regents of the University of California. The views and opinions of authors expressed herein do not necessarily state or reflect those of the United States Government or any agency thereof or the Regents of the University of California.

\vspace{2mm}
\noindent
This research used resources of the National Energy Research Scientific Computing Center (NERSC), a U.S. Department of Energy Office of Science User Facility located at Lawrence Berkeley National Laboratory, operated under Contract No. DE-AC02-05CH11231 using NERSC award m4055-ERCAP0020612.

\vspace{2mm}
\noindent
The authors thank Jeffrey Donatelli from Lawrence Berkeley National Laboratory for reviewing the manuscript. 

\paragraph{Author Contribution}
M.M.N. had the initial idea motivated by discussions with the the whole team, derived the mathematics for the kernels with help from M.D.R. and K.G.R., implemented the code with help from H.K., and wrote the first draft of the manuscript. M.D.R. revised the statistical parts of this work and the manuscript. K.G.R. checked and corrected the mathematical derivations. H.K. devised the technical components of the HPC implementation. The whole team revised the manuscript.

\bibliographystyle{plainnat}
\bibliography{literature}
\end{document}